\documentclass[conference]{IEEEURAI}
 \pdfoutput=1
\usepackage{graphicx}

\usepackage{epstopdf}
\usepackage{amsmath}
\usepackage{algorithm}
\usepackage{algorithmic}
\usepackage{gensymb}
\usepackage{url}
\usepackage{multirow}
\usepackage[table]{xcolor}
\usepackage{color, colortbl, soul} 
\usepackage[colorinlistoftodos]{todonotes}
\usepackage[binary-units=true]{siunitx}
\usepackage{subfigure}
\usepackage{dblfloatfix}

\DeclareGraphicsExtensions{.jpg, .png, .eps, .pdf}  

\begin{document}
\IEEEoverridecommandlockouts
\IEEEpubid{\makebox[\columnwidth]{\copyright 2017 IEEE \hfill} \hspace{\columnsep}\makebox[\columnwidth]{ }}
%
\title{Modelling the Influence of Cultural Information on Vision-Based Human Home Activity Recognition}

\author{\IEEEauthorblockN{Roberto Menicatti, Barbara Bruno and Antonio Sgorbissa}
\IEEEauthorblockA{Department of Informatics, Bioengineering, Robotics and System Engineering,\\University of Genova, Via Opera Pia 13, 16145 Genova, Italy\\(E-mail: roberto.menicatti@dibris.unige.it barbara.bruno@unige.it antonio.sgorbissa@unige.it)}}

\maketitle

\begin{abstract}
Daily life activities, such as eating and sleeping, are deeply influenced by a person's culture, hence generating differences in the way a same activity is performed by individuals belonging to different cultures. We argue that taking cultural information into account can improve the performance of systems for the automated recognition of human activities. We propose four different solutions to the problem and present a system which uses a Naive Bayes model to associate cultural information with semantic information extracted from still images. Preliminary experiments with a dataset of images of individuals lying on the floor, sleeping on a futon and sleeping on a bed suggest that: i) solutions explicitly taking cultural information into account are more accurate than culture-unaware solutions; and ii) the proposed system is a promising starting point for the development of culture-aware Human Activity Recognition methods.
\end{abstract}


%
\IEEEpeerreviewmaketitle

\section{Introduction}

All human beings are cultural beings. \textit{Culture} is defined as the shared way of life of a group of people, that includes beliefs, values, ideas, language, communication, norms and visibly expressed forms such as customs, art, music, clothing, food, and etiquette. Culture influences individuals' lifestyles, personal identity and their relationship with others both within and outside their culture \cite{Papadopoulos06}. Building on these premises, a recent research trend explores the influence of people's culture on their relationship with robots, aiming at assessing its impact on factors, such as acceptability and trust, which are of crucial importance for all applications of robots as personal assistants \cite{Trovato13,Eresha13,Joosse14b,Andrist15}.

In the context of Ambient Assisted Living (AAL), the recognition of human daily activities (and, in particular, of the Activities of Daily Living identified by gerontologists as tightly correlated with a person's autonomy \cite{Katz59,Lawton69}) is crucial to assess the health status of the assisted person. To this aim, vision-based Human Activity Recognition (HAR) systems are gaining more and more importance. Normally, HAR is performed on video streams rather than still images, as shown in some detailed surveys \cite{Chaquet2013, Ramanathan2014}. However, methods based on video streams usually only consider the human silhouette (through tracking or background subtraction) or are based on local small regions of interest (corresponding to different motion patterns) \cite{Poppe2010}, thus disregarding 
the additional relevant semantic information which can be deduced by 
the environment surrounding the person performing the action. 

It is an established fact that Activities of Daily Living, e.g. \textit{eating} and \textit{sleeping}, are carried out in different ways and different places of the house, in accordance with the cultural identity of the person \cite{Meghani-Whise1996}. Mulholland and Wyss \cite{Mulholland2001} report that, for example, in many parts of Asia, postures such as squatting, kneeling or sitting cross-legged on the floor are more common than using a chair. In Japan a kneeling posture is commonly adopted to perform daily activities such as eating, socializing, and religious or traditional ceremonies such as the tea ceremony. In Asia and the Middle East people sit cross-legged on mats and tatami for resting, socializing, eating, working, or leisure or spiritual activities such as yoga. 

Conversely, postures assuming a direct contact with the floor are generally uncommon in European countries and are often associated with potentially dangerous situations (such as a sudden illness, fall, or faint).

Since culture influences and pervades most of the actions of a person, and particularly everyday activities such as eating, sleeping and toileting, we argue that in-home assistive robots should not only be \textit{culture-aware} when directly interacting with a person, but also, and more generally, able of evaluating any type of user-related information in light of said person's culture and preferences. 

As a preliminary step towards the development of culture-aware HAR systems, we address the problem of: i) determining whether a person is sleeping or lying in a potentially dangerous situation; ii) taking into account the influence of culture on the way in which the \textit{sleeping} activity is performed, in particular by considering the case in which the person sleeps on a bed, as it is common in European countries, and the case in which the person sleeps on a futon, as it is common in Japan. More precisely, the contribution of the article is two-fold: i) the enhancing of the Cloud-based HAR framework presented in \cite{Menicatti2017} to include cultural information and increase the chances of a right classification; and ii) the comparison of the results obtained by taking culture into account at different levels for the (vision-based) recognition of a person who is sleeping or lying in a potentially dangerous situation.

The article is organized as follows. Section \ref{sec: Problem Statement} outlines the problem statement and proposes four different solutions for embedding cultural information in the process of recognizing daily activities. Section \ref{sec:Method} describes the method we propose for including cultural information both in the training and in the testing phase of a vision-based HAR system. Section \ref{sec: Tests and Results} compares the tests performed and the results obtained by using the different solutions adopted. Conclusions follow.

\section{Problem Statement}
\label{sec: Problem Statement}

Human Activity Recognition systems based on visual information usually require the execution of two distinct phases: during the \textit{training} phase, a number of examples (i.e., labelled images) of the activity to recognize are used for the creation of its model; then, during the \textit{testing} phase, an unlabelled recording (i.e., a new image) is analysed in light of the available models, and labelled as an instance of the one that better matches it. In this context, it is possible to envision different solutions for modelling and recognizing activities in which cultural factors play a non-negligible role.

\begin{enumerate}
\item \textit{Individual-specific}. Trivially, if all examples used in the training phase, as well as all the recordings used in the testing phase, belong to one and the same person, the system will always be aligned with that person's culture. This solution, which best captures the unique cultural traits of an individual, requires a long set up and does not allow for exploiting similarities among different persons.
\item \textit{Culture-unaware}. At the opposite end of the spectrum with respect to the individual-specific solution, culture-agnostic systems rely on a large number of examples, from many different individuals, for the creation of models general enough to be valid for different cultures. For example, in the case of the \textit{sleeping} activity, mixing examples from Japanese and European individuals in the training phase might lead to the creation of a model which does not rely on the presence of a bed. This solution minimizes the set up time, since the training is done only once, but, arguably, at the expenses of a reduced accuracy in capturing person-specific traits.
\item \textit{Culture-aware training}. A more interesting solution envisions the creation of culture-specific models of all activities in which cultural factors may play a non-negligible role, thus leading, for example, to the creation of two models of the \textit{sleeping} activity, e.g., \textit{sleeping-futon} and \textit{sleeping-bed}, respectively for the Japanese and European culture. In the testing phase, recordings of a European person sleeping are likely to better match the \textit{sleeping-bed} model, while recordings of a Japanese person sleeping are likely to be more often labelled as occurrences of the \textit{sleeping-futon} activity. This solution builds on the assumption that it is possible to achieve a good trade-off between recognition accuracy and generality of the models by taking cultural differences into account explicitly during the training phase and implicitly in the testing phase, when the system relies on sensory cues only to infer the culture of the person.
\item \textit{Culture-aware training and testing}. Let us consider the image of a person lying on the floor. In the absence of explicit information about his/her cultural profile, the HAR system might lack clear evidence to discriminate between a Japanese person sleeping on a thin futon and a European person who fell and is in need of assistance. This solution builds on the assumption that, by explicitly considering cultural information during both the training and the testing phase, it is possible to improve the system accuracy, with no loss in the generality of the models.
\end{enumerate}

The first solution, which does not specifically address the problem of modelling the influence of culture on daily activities, is not considered in this article. The interested reader might find relevant information about individual-specific HAR systems in \cite{Menicatti2017}.

The second and third solutions allow for the adoption of any vision-based HAR method, since the cultural information is encoded in the images collected for the training phase, and, in the third solution, explicitly expressed by creating different, culture-dependent, models of the same activity.

Conversely, to the best of our knowledge, there is no vision-based HAR system allowing for explicitly considering cultural information during the testing phase. The following Section outlines the method we propose to this aim, together with the rationale for the design choices supporting it.

\section{Method}
\label{sec:Method}

Albeit scarce, it is possible to find in the literature examples of robotic systems which model cultural information to tune their behaviour towards an individual. Torta et al. propose a method to parametrize the interpersonal distance and direction of approach that the robot should use when talking to a person \cite{Torta11}. Information about the acceptability of different values of distance and orientation is encoded in a multi-dimensional function and combined with contingent sensory information (for example, concerning the presence of obstacles) in a Bayesian inference mechanism with a particle filter to identify a suitable target pose for the robot.

A more complex example describes a framework for the learning and selection of culturally appropriate greeting gestures and words \cite{Trovato15b}. In the proposed system, an initial set of gestures and words is extracted from video and text corpora, and initial associations between gestures and words and a number of cultural factors of relevance are taken from literature in social studies and expressed as conditional probabilities in a Naive Bayes classifier. At run-time, the user's cultural profile in terms of the cultural factors is computed and used to identify the greeting gestures and words which better match the profile.

In accordance with literature findings, we propose the use of a Naive Bayes classifier for associating cultural information to visual features.

\begin{figure}[t]
	\centering
	\subfigure[Training]{
		\includegraphics[width=1\linewidth]{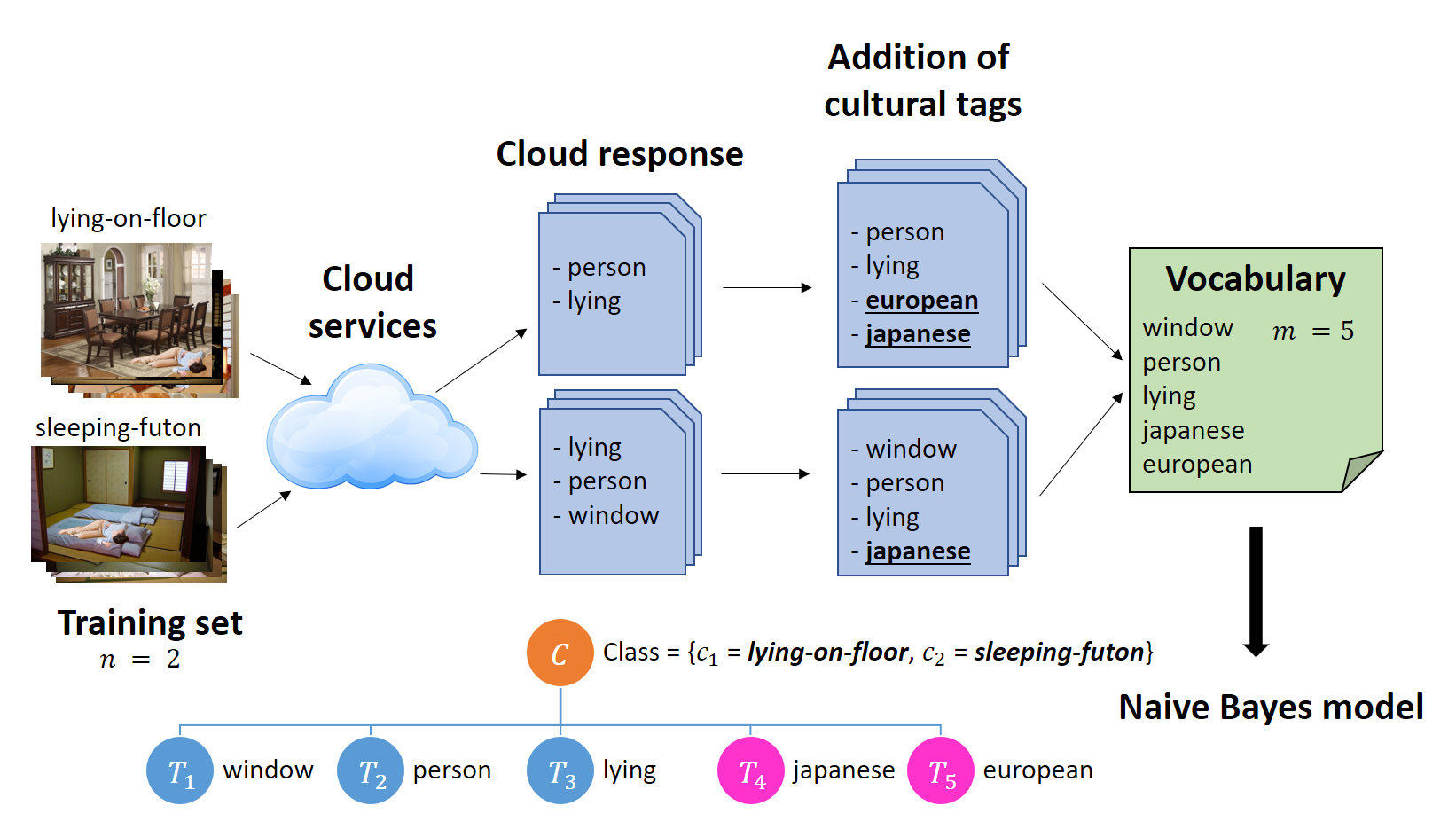}
		\label{fig: Training}
	}  						
	
	\subfigure[Testing]{
		\includegraphics[width=1\linewidth]{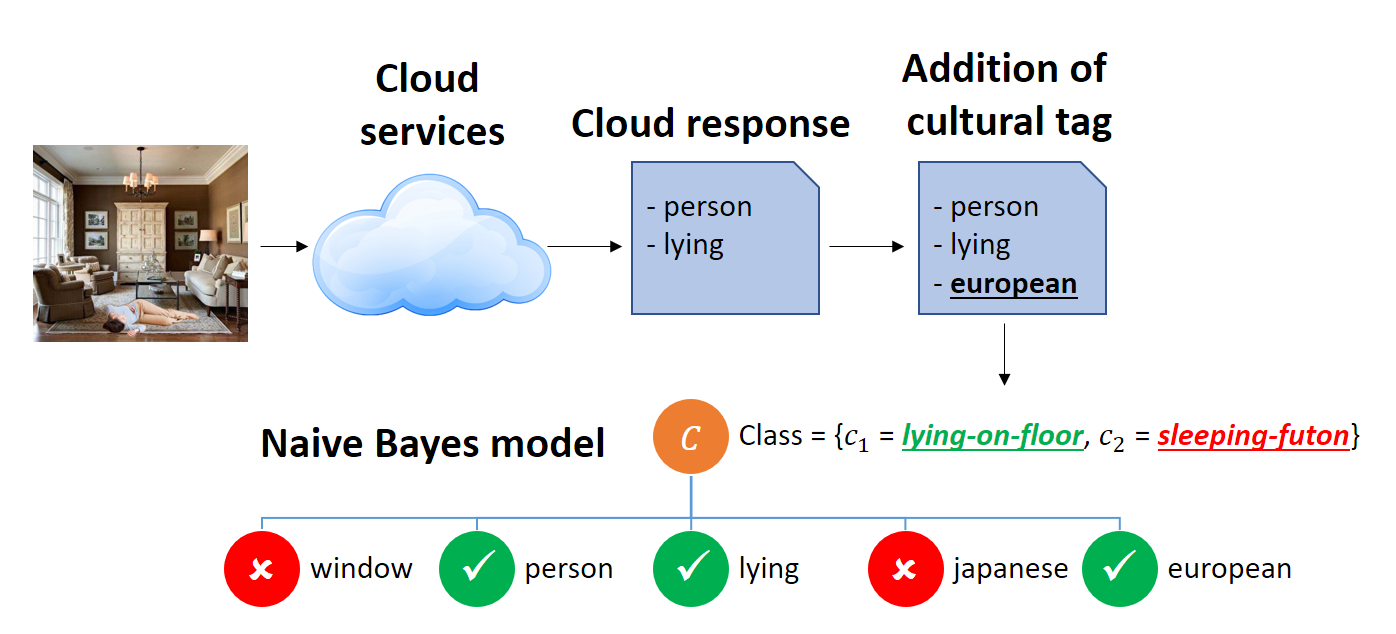}
		\label{fig: Testing}
	}
	\caption[]{Architecture of the training/testing phase of the proposed HAR system with culture aware training and testing
	\label{fig: Architecture}}
\end{figure}

In particular, in this work we rely on the Cloud-based HAR (CHAR) framework presented in \cite{Menicatti2017} and extend it to include cultural information. CHAR exploits the computer vision cloud services provided by Clarifai\footnote{\url{http://www.clarifai.com/}}, Microsoft\footnote{\url{https://azure.microsoft.com/it-it/services/cognitive-services/computer-vision/}} and Google\footnote{\url{https://cloud.google.com/vision/}} to extract semantic information from static images. In particular, the three cloud services return a list of tags describing the objects, the environments, the actions etc. detected in the image without giving any information about their spatial relation. As Fig. \ref{fig: Training} shows, during the training phase a number of training images for each activity of interest are available. The tags returned by the three cloud services for all the training images are collected and used to train a Naive Bayes model, in which the parent node represents the activity (henceforth also referred to as the \textit{class}) and each child node is a tag extracted from the training sets. An activity is therefore modelled as a probability distribution over the whole set of possible tags. During the testing phase (see Fig. \ref{fig: Testing}), an image is given to the cloud services to retrieve the associated tags, which are then compared with the available models to identify the activity more likely to be represented in the image.

A number of reasons support the choice of this method as a starting point for the development of a HAR framework with culture-aware training and testing: i) by relying on a Naive Bayes model and semantic tags, it is very close to the aforementioned methods proposed in the literature for the modelling of cultural factors; ii) it can be easily adapted to match the requirements of the \textit{culture-unaware} and \textit{culture-aware training} solutions, to allow for a meaningful comparison of the different approaches; iii) it is based on publicly accessible online services, thus enforcing reproducibility.
Figure \ref{fig: Training} on the right side shows our proposal for enhancing the CHAR system with cultural information. In particular, in the training phase to the list of tags returned by the cloud services for each training image, we add a tag describing the cultural identity of the person depicted in the image, assuming that this information is available for all the images in the training set. Let us refer to this kind of tag as \textit{cultural tag}. The vocabulary will then include the cultural tags, thus extending the Naive Bayes model with additional children \textit{cultural nodes} (shown in fuchsia in Fig. \ref{fig: Training}), one for each cultural tag included.

As for the other tags, the cultural tags have a different probability distribution for each trained class. If a class represents a culture-specific activity or situation (e.g. sleeping on a futon), the probability value of the corresponding cultural tag (e.g. a tag \textit{japanese}) for that class is equal to $1$ while all the other cultural tags, representing other cultures, (e.g. tags as \textit{italian}, \textit{mexican}...) have probability $0$. If, instead, the class is culture-independent because the scene represented is an activity which is performed in the same way in different countries the probability distribution of the cultural tags will be proportional to the measure of the presence of each culture considered in the training class. Consider, for example, a class \textit{reading} which contains $6$ images of different persons reading. In $3$ images the person is Italian, in $1$ is Japanese and in $2$ images is Mexican. The probability of the corresponding cultural tags \textit{italian}, \textit{japanese} and \textit{mexican} for that class are $3/6$, $1/6$ and $2/6$ respectively. 

In the testing phase (Fig. \ref{fig: Testing}) a tag with the cultural profile of the person shown in the tested image is added to the list of tags returned by the cloud services. Also in this case it is safe to assume that such cultural knowledge is available, since a device performing image recognition in an AAL scenario (e.g. an assistive robot) can be given the knowledge of the cultural identity of the user during its setup. The presence of a cultural tag and the absence of the other ones will set the states of the relative cultural nodes, thus influencing the classification of the image.

\section{Tests and Results} 
\label{sec: Tests and Results}

As anticipated in Section \ref{sec: Problem Statement}, in this work we focus only on the second, third and fourth solutions described, that we denote as:
\begin{itemize}
	\item \textit{Culture-Unaware} (CU);
	\item \textit{Culture-Aware Training} (CAT);
	\item \textit{Culture-Aware Training and Testing} (CATT).
\end{itemize}

To address the problem of determining whether a person is sleeping or lying in a potentially dangerous situation, taking cultural information into account, we consider the following three situations: 
\begin{enumerate}
	\item sleeping on a bed (associated to European culture);
	\item sleeping on a futon (associated to Japanese culture);
	\item lying on the floor (not associated to any specific culture).
\end{enumerate}
Discriminating one of such situations from the others is not trivial: all such activities involve a person in the same posture (i.e., lying) and two of them ("sleeping on a futon" and "lying on the floor") are strikingly similar. A wrong classification could, for example, classify a person who is sleeping on a futon as someone who has fallen and, in the context of AAL, result in a false alarm raised by the robot, or monitoring system, and thus a reduced reliability. The opposite misclassification has even worse consequences: a robot might fail to alert of a fall, by wrongly classifying a person lying on the floor as one sleeping on a futon. 

\begin{figure*}[thpb]
	\centering
	\includegraphics[width=1\linewidth]{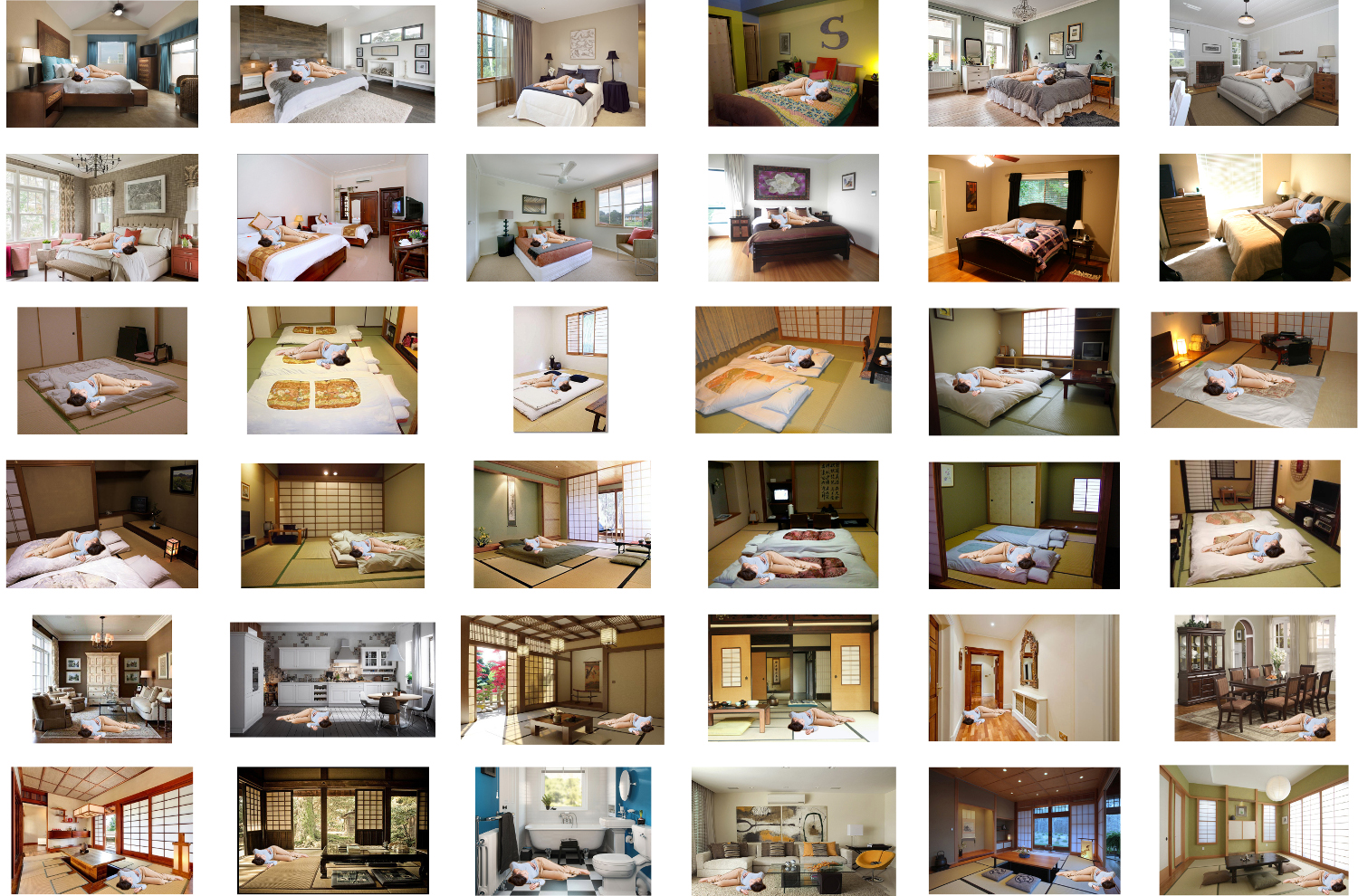}
	\caption{Dataset: subclass \textit{sleeping-bed} (rows 1-2), subclass \textit{sleeping-futon} (rows 3-4), class \textit{lying-on-floor} (rows 5-6)}
	\label{fig: Dataset2}
\end{figure*}

All of our tests have been performed offline; in particular, we have collected a dataset of $36$ images (Fig. \ref{fig: Dataset2}), divided in: 
\begin{itemize}
	\item class \textit{sleeping}, $24$ images divided in:
	\begin{itemize}
		\item subclass \textit{sleeping-bed}, $12$ images;
		\item subclass \textit{sleeping-futon}, $12$ images;
	\end{itemize}
	\item class \textit{lying-on-floor}, $12$ images.
\end{itemize} 

To further increase the ambiguity, all the images were obtained by putting the same figure of a lying-down person over different backgrounds, thus visually simulating the three different cases. While the images of subclasses \textit{sleeping-bed} and \textit{sleeping-futon} have respectively European and Japanese home backgrounds only, for the \textit{lying-on-floor} class we have collected $6$ images with a European home background and $6$ images with a Japanese home background.

\subsection{Culture-Unaware}
\label{subsec: Culture-Unaware}
As defined in Section \ref{sec: Problem Statement}, a culture-unaware solution does not take into explicit consideration the different ways in which a same activity is performed according to different cultures. Therefore, in this case we consider a training dataset composed of 12 images for each class (i.e., "\textit{lying-on-floor}" and "\textit{sleeping}"). The training set of the \textit{sleeping} class is equally split among its two subclasses.

We have adopted the standard CHAR system for the recognition, and used \textit{k-fold cross validation} upon the dataset for the training and testing. We have randomly divided each class in $3$ subsets of $4$ images each, then, by taking one subset of each class for the testing set and the remaining two subsets of each class for the training set, we have combined them in $9$ possible combinations. Therefore, each subset is used $3$ times for the testing and $6$ times for the training.

\subsection{Culture-Aware Training}
\label{subsec: Culture-Aware Training}
The CAT solution assumes an explicit modelling of cultural information in the training phase only. In this case we use the full dataset, divided in the three classes mentioned before. During the training, the cultural information is taken into account by considering the \textit{sleeping} class as made of the two, independent subclasses \textit{sleeping-bed} and \textit{sleeping-futon} which allows for the separation of the different ways the act of sleeping is performed and the different ways this situation is depicted in the images.

As for the CU solution, also in this case we have used the standard CHAR system and k-fold cross validation. The $3$ subsets of each class have been combined in $27$ folds using, at each fold, one subset for testing and two for training for each class. Each subset is used $9$ times for the testing and $18$ times for the training. All subsets of the \textit{lying-on-floor} class contain two images with a "European" background and two images with a "Japanese" background.

\subsection{Culture-Aware Training and Testing}
\label{subsec: Culture-Aware Training and Testing}
The CATT solution assumes an explicit modelling of cultural information not only in the training phase but also at the moment of classifying an image. We have again used k-fold cross validation, with the same folds of the CAT solution. However, instead of using CHAR, we have used its modified version, described in Section \ref{sec:Method}. The modified version of CHAR has been implemented for the training and testing of this specific case as follows.

\subsubsection{Training}
\label{subsubsec: Training}
The training set of each class consists of $8$ images. In accordance with the method explained in Section \ref{sec:Method}, we have added a cultural tag to all the training images. In particular, we have added the tag:
\begin{itemize}
	\item \textit{european} to all the images of the class \textit{sleeping-bed} and to the images of the class \textit{lying-on-floor} with the "European" background;
	\item \textit{japanese} to all the images of the class \textit{sleeping-futon} and to the images of the class \textit{lying-on-floor} with the "Japanese" background.
\end{itemize}

The probability distribution of the two cultural tags is shown in Table \ref{tab: probabilities}. Please notice that the values in the table are representative of how we have built the training set for the purpose of assessing the performance of the solution and do not represent any real probability distribution of these cultural factors.

The two cultural tags and their probabilities are automatically included in the training of the Naive Bayes model.

\begin{table}[]
\centering
\caption{Probability distribution of the additional cultural tags \textit{european} and \textit{japanese}}
\label{tab: probabilities}
\begin{tabular}{l|c|c||c|c|}
	\cline{2-5}
	                                      &           \multicolumn{2}{c||}{\textit{european}}           &           \multicolumn{2}{c|}{\textit{japanese}}           \\ \cline{2-5}
	                                      & \multicolumn{1}{l|}{present} & \multicolumn{1}{l||}{absent} & \multicolumn{1}{l|}{present} & \multicolumn{1}{l|}{absent} \\ \hline
	\multicolumn{1}{|l||}{sleeping-bed}   &              1               &              0               &              0               &              1              \\ \hline
	\multicolumn{1}{|l||}{sleeping-futon} &              0               &              1               &              1               &              0              \\ \hline
	\multicolumn{1}{|l||}{lying-on-floor} &             0.5              &             0.5              &             0.5              &             0.5             \\ \hline
\end{tabular}
\end{table}

\subsubsection{Testing}
\label{subsubsec: Testing}
The testing set of a class consists of 4 images. The two cultural tags have been added in the same way as for the training.

The size of the training set (8 pictures) and testing set (4 pictures) of a class is the same for all the methods. This, of course, leads to a different overall number of test samples for the three solutions when using k-fold cross validation, since the number of classes is 2 in the CU case and 3 in the CAT and CATT ones. However, it allows for a better comparison of the performance of the three methods over each single class of image.

\begin{figure*}[thpb]
	\centering
	\subfigure[CU]{
			\includegraphics[width=0.31\linewidth]{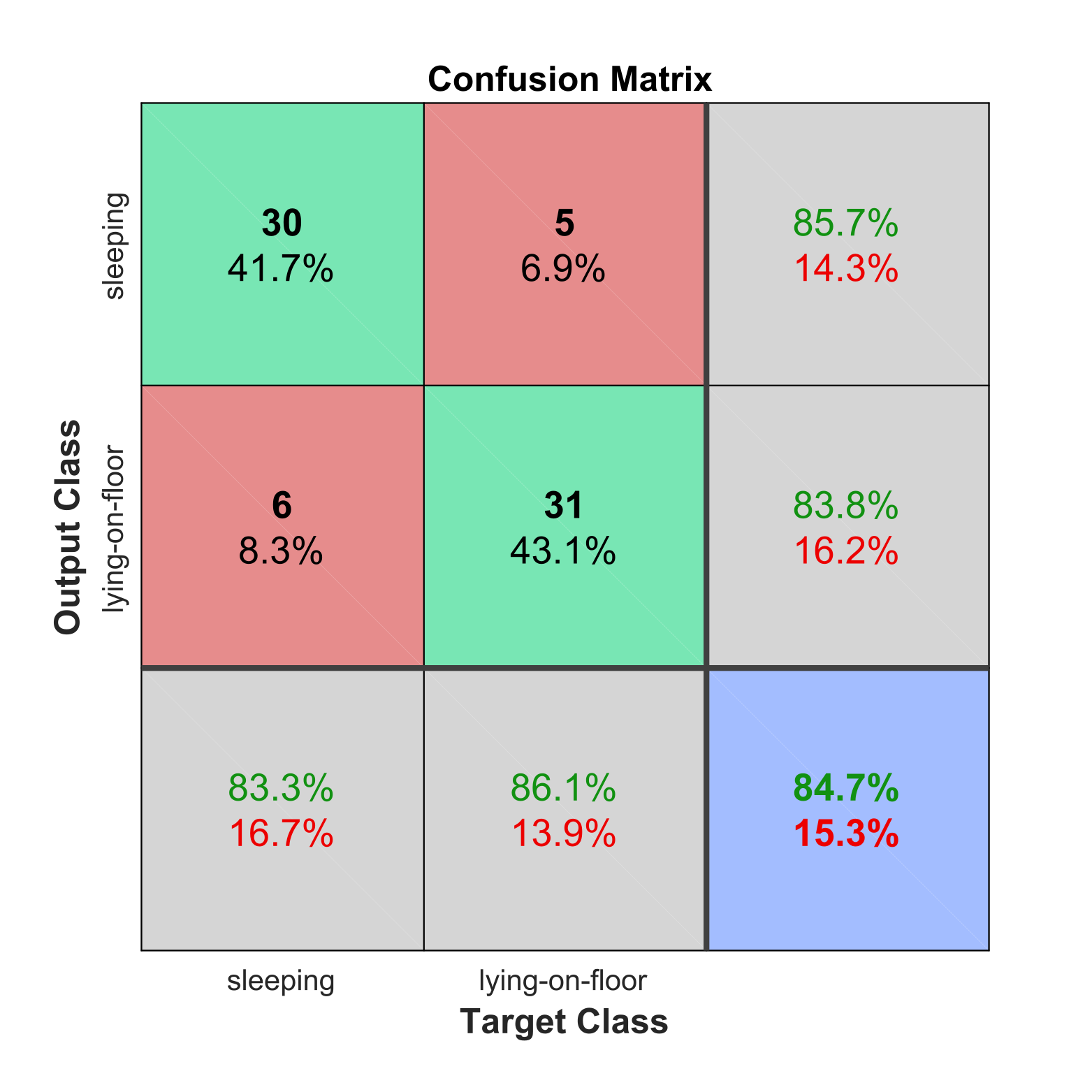}
			\label{fig: confusionMatrixTwoClasses}
	} 
	\subfigure[CAT]{
		\includegraphics[width=0.31\linewidth]{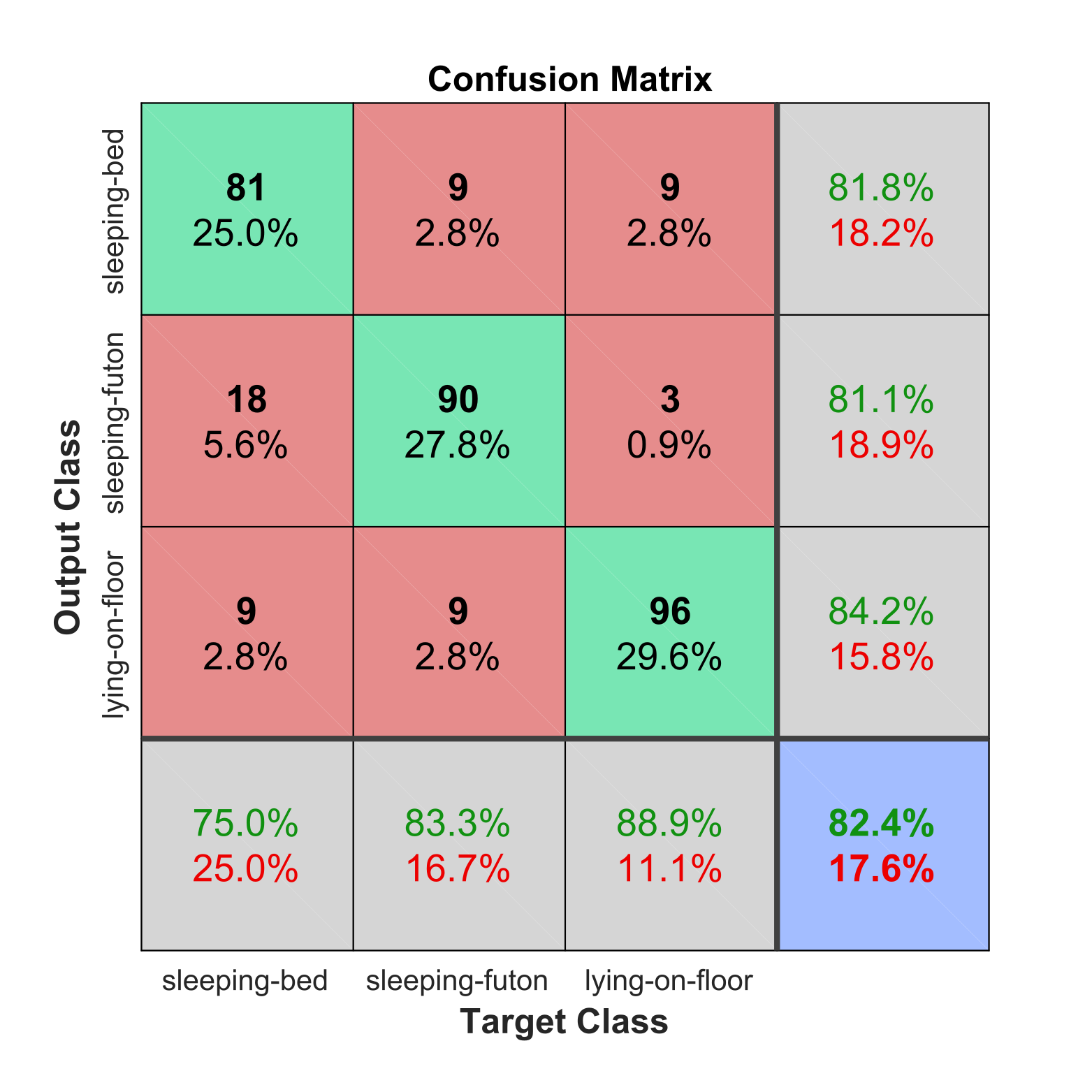}
		\label{fig: confusionMatrixNoTag}
	}  						
	\subfigure[CATT]{
		\includegraphics[width=0.31\linewidth]{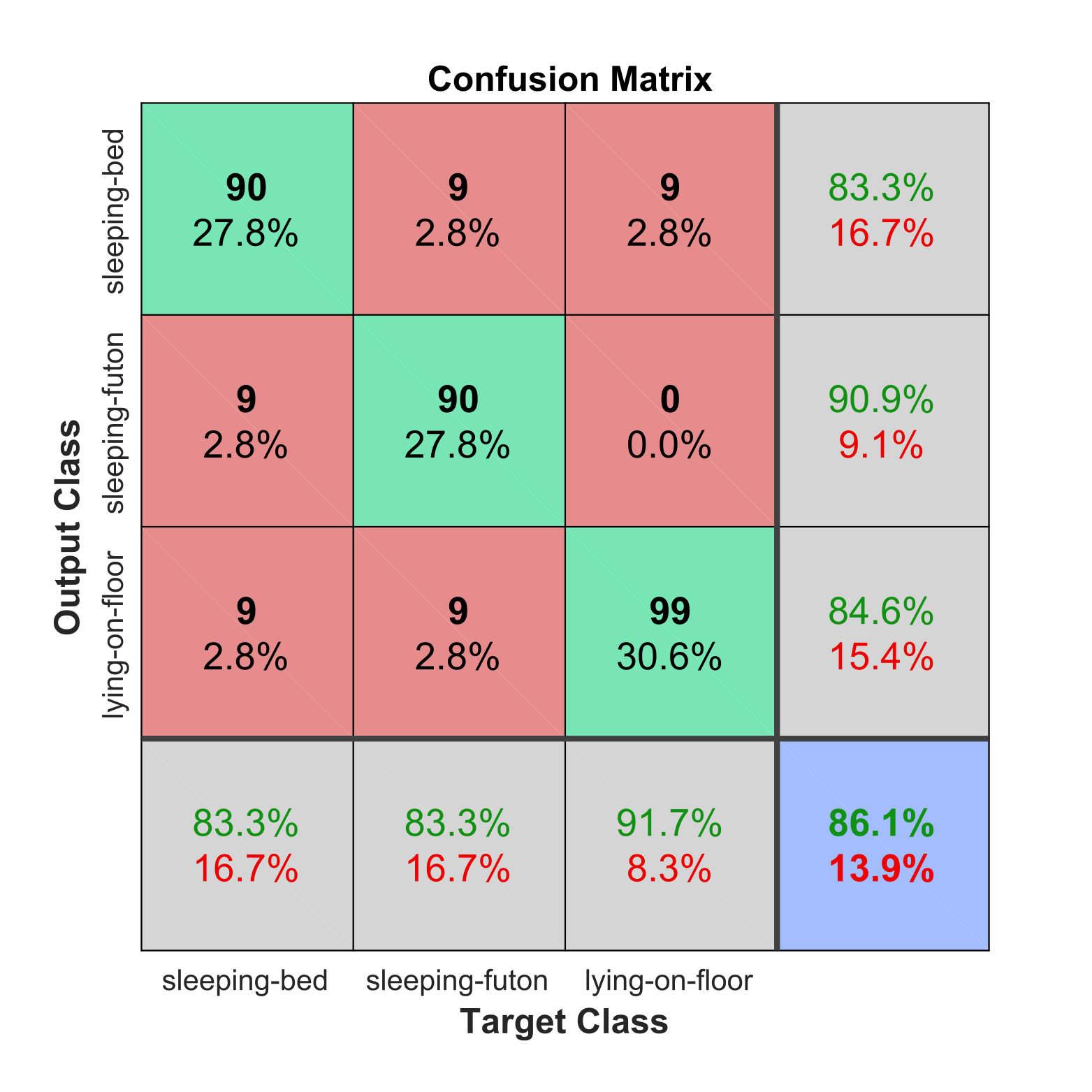}
		\label{fig: confusionMatrixWithTag}
	}
	\caption[]{Confusion matrices for CU (a), CAT (b), CATT(c) solutions
	\label{fig: confusionMatrices}}
\end{figure*}
\subsection{Results}
\label{sec: Results}
In order to evaluate the performance of the three different solutions we have computed their confusion matrices, shown in Fig. \ref{fig: confusionMatrices}. With reference to the figure, the columns correspond to the actual class of the tested images (target class) and the rows represent the predicted class (output class). The diagonal cells (green background) show the number of True Positives of each class, i.e. the number of images which have been classified correctly, and the percentage over the overall number of images in all the testing sets. The off diagonal cells show, instead, the number of wrong detections. The last row shows the \textit{recall} (or true positive rate) of each class, the last column shows the \textit{precision} of each class, the bottom-right cell (blue background) shows the overall accuracy.

By comparing the two confusion matrices for solutions CAT and CATT we can see that $9$ \textit{sleeping-bed} images which are wrongly classified as \textit{sleeping-futon} in the case without cultural tags are, instead classified correctly when these tags are included (cells of first and second row, first column of Fig. \ref{fig: confusionMatrixNoTag} and Fig. \ref{fig: confusionMatrixWithTag}). Moreover, the $3$ \textit{lying-on-floor} images wrongly classified as \textit{sleeping-futon} in the first case, are then classified correctly when including the cultural tags (cells of second and third row, third column of Fig. \ref{fig: confusionMatrixNoTag} and Fig. \ref{fig: confusionMatrixWithTag}).

Moreover, for those images which both CAT and CATT solutions have classified incorrectly, we have computed the difference in the confidence scores and averaged over the different images. We have noticed that the CATT solution presents an average confidence score for the misclassified images lower by $5\%$ than the one of the CAT solution.

In order to better compare the performance of CAT and CATT with CU we have grouped the results of the subclasses \textit{sleeping-bed} and \textit{sleeping-futon} for the case CAT and CATT as if they were a single class \textit{sleeping} and computed the aggregate recall and precision values. With reference to the confusion matrices of Fig. \ref{fig: confusionMatrixNoTag} and Fig. \ref{fig: confusionMatrixWithTag}, let us call $M$ indifferently one of the two matrices, $i$ the row index and $j$ the column index. Concretely, for the superclass \textit{sleeping} both in the case of CAT and CATT we have computed:
\begin{equation}
	recall_{s} = \frac{TP_s}{TP_s + FN_s} = \frac{\displaystyle\sum_{i=1}^{2}\sum_{j=1}^{2}M_{ij}}{\displaystyle\sum_{i=1}^{2}\sum_{j=1}^{2}M_{ij} + \sum_{j=1}^{2}M_{3j}}
\end{equation}
\begin{equation}
	precision_{s} = \frac{TP_s}{TP_s + FP_s} = \frac{\displaystyle\sum_{i=1}^{2}\sum_{j=1}^{2}M_{ij}}{\displaystyle\sum_{i=1}^{2}\sum_{j=1}^{2}M_{ij} + \sum_{i=1}^{2}M_{i3}}
\end{equation}
where \textit{TP} stands for \textit{true positives}, \textit{FN} stands for \textit{false negatives}, \textit{FP} stands for \textit{false positives} and the subscript \textit{s} stands for the class \textit{sleeping}. 

\begin{figure}[t]
	\centering
	\subfigure[Recall]{
		\includegraphics[width=1\linewidth]{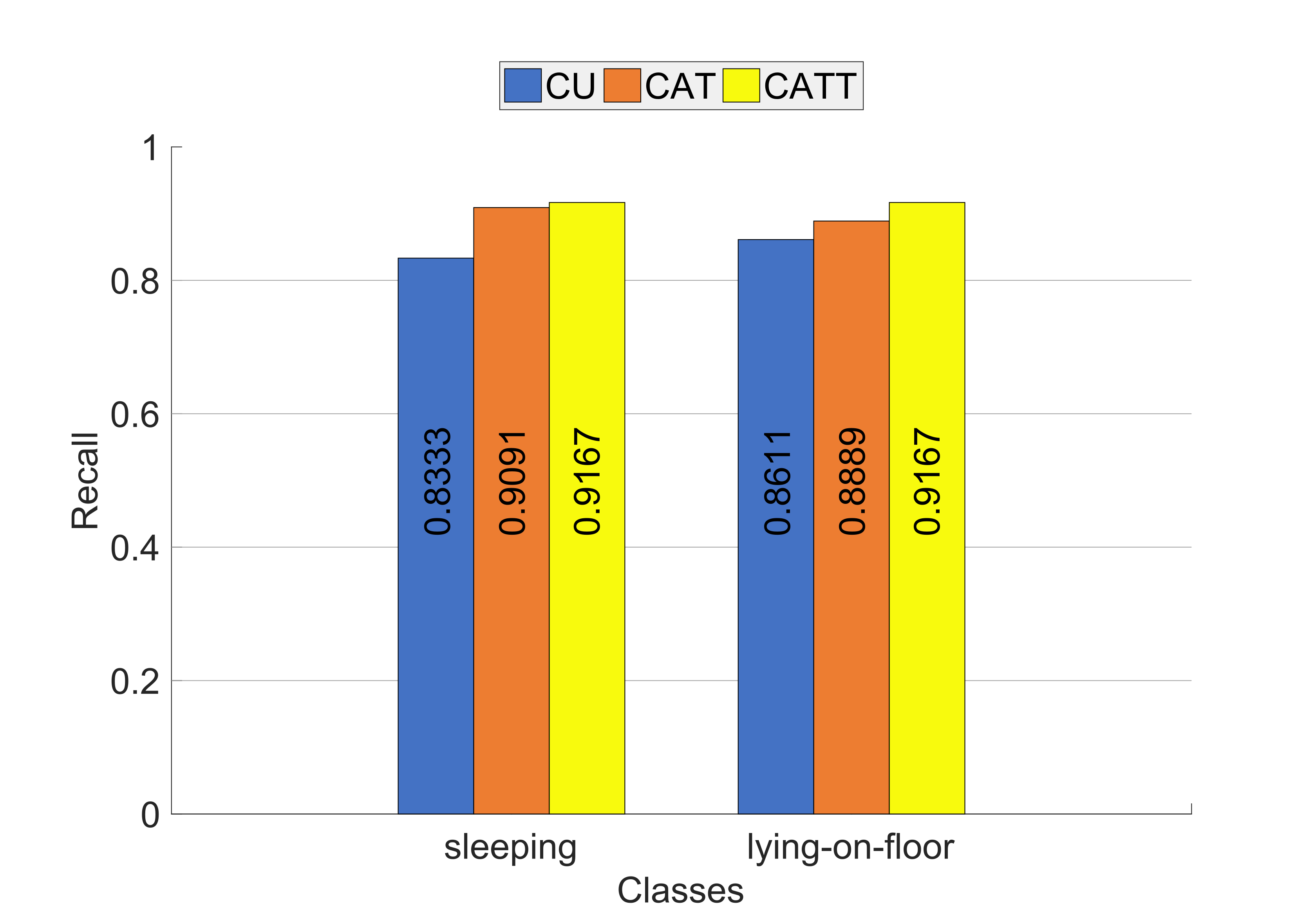}
		\label{fig: recall}
	}  						
	
	\subfigure[Precision]{
		\includegraphics[width=1\linewidth]{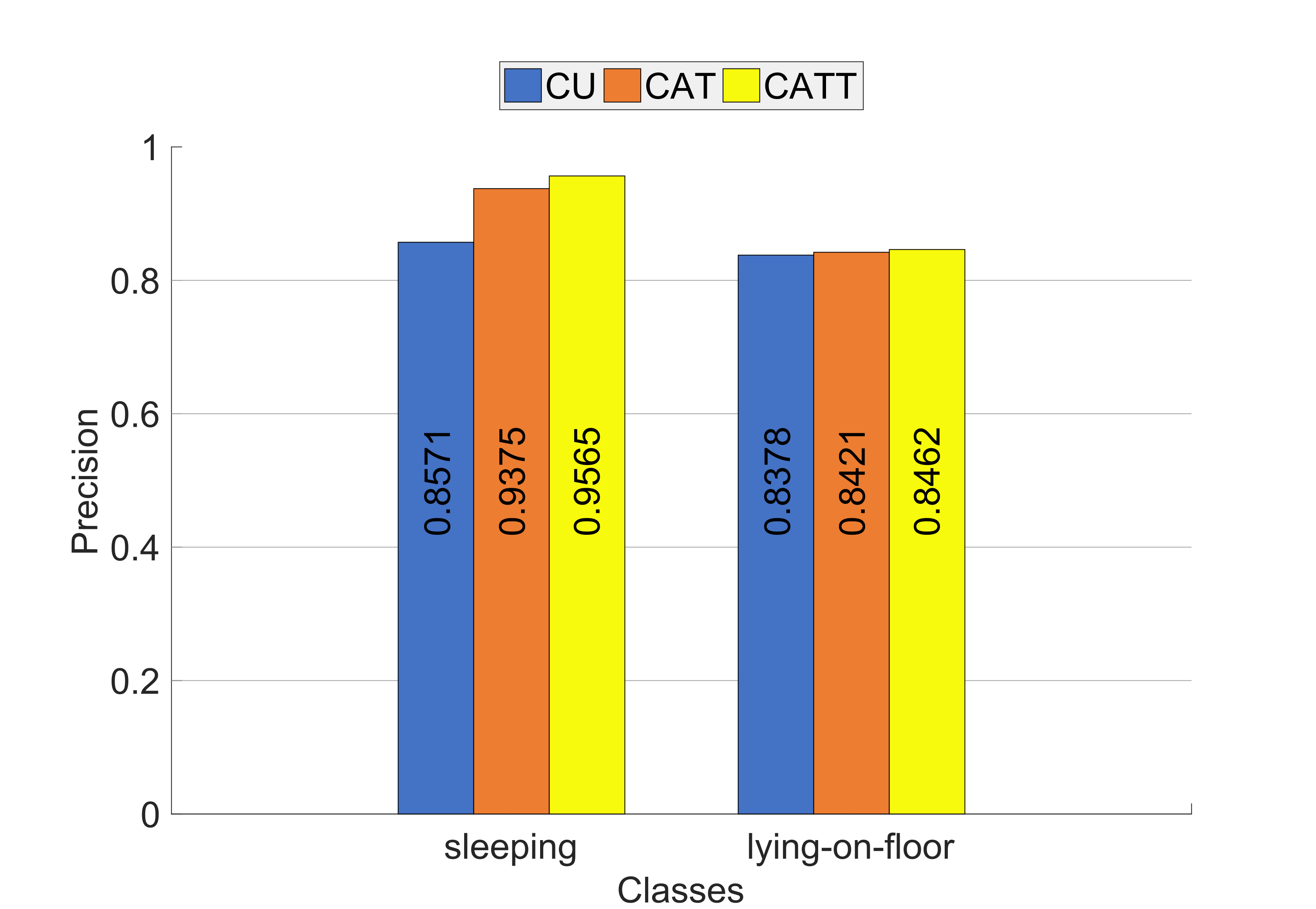}
		\label{fig: precision}
	}
	\caption[]{Comparison of recall (a) and precision (b) between the three solutions (CU, CAT, CATT)
	\label{fig: recallprecision}}
\end{figure}

Finally, the plots in Fig. \ref{fig: recallprecision} compare, for each class, the recall and the precision obtained through the three different solutions adopted. As expected, the figures show that adding cultural information improves the recognition performance, both in terms of precision and recall, with the CATT solution being the most accurate.
\addtolength{\textheight}{-11cm}   

\section{Conclusions}
\label{sec: Conclusions}

In this article, we have discussed the influence of a person's culture on his daily activities, and, as a consequence, the need for taking cultural information into account when designing automated Human Activity Recognition systems.

In the field of vision-based HAR systems for the recognition of Activities of Daily Living, we have identified three possible non-trivial solutions to the problem, and compared their performance. Of these three solutions, one attempts to create general models by involving examples from people belonging to different cultures in the same training class (CU), while the other two explicitly include cultural clues either in the training (CAT) or both in the training and testing phase (CATT).


As a possible implementation of the last approach, we have adopted a system for the recognition of daily activities and indoor scenes which relies on cloud-based computer vision services and a Naive Bayes model to represent activities as distributions of probabilities over a set of tags, and enhanced it with tags encoding relevant cultural information.

To compare the performance of the CU, CAT and CATT solutions, we have collected a dataset of images pertaining to two activities (i.e., sleeping and lying on the floor as a consequence of a fall, or a sudden illness), and distinguished between Japanese people sleeping on a futon, and European people sleeping on a bed.

Experiments support our hypothesis that: i) explicitly taking cultural information into account allows for a more accurate recognition; ii) the proposed method is a suitable choice as a culture-aware HAR system, with an overall precision above $84\%$ and an overall recall above $91\%$.

\section{Acknowledgement}
This work has been partly supported by the European Commission Horizon2020 Research and Innovation Programme under grant agreement No. 737858 (CARESSES).





\bibliographystyle{ieeetran}

\bibliography{bibliografia}

\end{document}